# Parallel processor scheduling: formulation as multi-objective linguistic optimization and solution using Perceptual Reasoning based methodology

Prashant K Gupta, *Member IEEE*, Pranab Kumar Muhuri, *Member IEEE*

*Abstract*- **In the era of Industry 4.0, the focus is on the minimization of human element and maximizing the automation in almost all the industrial and manufacturing establishments. These establishments contain numerous processing systems, which can execute a number of tasks, in parallel with minimum number of human beings. This parallel execution of tasks is done in accordance to a scheduling policy. However, the minimization of human element beyond a certain point is difficult. In fact, the expertise and experience of a group of humans, called the experts, becomes imminent to design a fruitful scheduling policy. The aim of the scheduling policy is to achieve the optimal value of an objective, like production time, cost, etc. In real-life situations, there are more often than not, multiple objectives in any parallel processing scenario. Furthermore, the experts generally provide their opinions, about various scheduling criteria (pertaining to the scheduling policies) in linguistic terms or words. Word semantics are best modelled using fuzzy sets (FSs). Thus, all these factors have motivated us to model the parallel processing scenario as a multi-objective linguistic optimization problem (MOLOP) and use the novel perceptual reasoning (PR) based methodology for solving it. We have also compared the results of the PR based solution methodology with those obtained from the 2-tuple based solution methodology. PR based solution methodology offers three main advantages viz., it generates unique recommendations, here the linguistic recommendations match a codebook word, and also the word model comes before the word. 2-tuple based solution methodology fails to give all these advantages. Thus, we feel that our work is novel and will provide directions for the future research.**

*Index Terms*- **Computing with words, Hao-Mendel approach (HMA), Industry 4.0, Interval approach (IA), Parallel processor scheduling, Perceptual computing, Perceptual reasoning.**

## I. INTRODUCTION

In a number of activities such as business integrations, competitions, the requirement for product mix, etc., different types of computing systems are employed. These systems are powered by processors having similar or distinct processing capabilities and perform numerous tasks which can also be identical or distinct. These tasks are allocated to the processors and executed as independent tasks in parallel. This allocation and execution of tasks is called parallel processor scheduling. The tasks scheduled for the parallel processor scheduling need to be allocated resources such as memory, network bandwidth, etc. to ensure their timely execution. All the allocations are done by a suitable allocation policy commonly called the scheduling algorithm.

The scheduling concepts are quite useful in the scenario of Industry 4.0 [1-7]. Industry 4.0 aims to increase automation and reduce involvement of human beings to achieve faster production. However, experts are the class of human beings, whose opinions are very important while designing successful systems [14], [30]. In fact, it's not possible to get maximum benefit in the system design without the expert knowledge. However, experts tend to express their opinions naturally using linguistic terms or words. These opinions may also pertain to values of various scheduling parameters, corresponding to the scheduling algorithm adopted by the system. As these values are expressed linguistically, therefore, they can be uncertain.

An attempt to design a scheduling policy based on the linguistic values of the scheduling parameters was made in [8]. Here, the authors presented the scenario of an automotive subcontract company, where a fixed number of welding jobs were accomplished by a given number of welders. A suitable scheduling policy was used to allocate the jobs to these welders. The authors modelled various scheduling factors as linguistic variables and represented their semantics using fuzzy sets (FSs) [9] (as various scheduling factors were uncertain and imprecise). The relationship between these linguistic variables was defined using if-then rules and Mamdani inference method was used to calculate the crisp values of each welder's processing times. The values of the times taken to weld their respective jobs by individual welders were called respective operation times, which were summed up and this quantity was called the overall completion time. The objective was to minimize the overall completion time.

The approach of [8] was undoubtedly an improvement of the previous works; however, it suffered from a limitation that the value of the overall completion time was generated in the numeric form. Since the values of scheduling factor were in linguistic form, the solution should also be in linguistic form because human beings understand linguistic information naturally. Therefore, recently a work was proposed by Gupta and Muhuri [10], which tried to overcome this limitation. The authors used the novel 2-tuple computing with words (CWW) approach for linguistic operation time computation, in the parallel processor scheduling scenario of [8]. The generation of linguistic operation time by [10], for parallel processor scheduling is quite useful for human beings.

However, the scheduling scenarios seldom have a single objective. Any organization employing the scheduling policy for a number of tasks will generally have multiple objectives. For example, any organization will be interested in minimizing the overall task completion time as well as maximizing the overall profit. This type of scenario can be

Manuscript submitted March, 2019
Prashant K Gupta and Pranab Kumar Muhuri are with Department of Computer Science, South Asian University, New Delhi, India.
(e-mail: guptaprashant1986@gmail.com, pranabmuhuri@cs.sau.ac.in)

modelled as a multi-objective linguistic optimization problem (MOLOP), where the two objectives are overall completion time and overall profit (We will discuss the details in Section III). The 2-tuple based solution methodology for such scenarios can be employed, for linguistic value computation of respective objectives. A 2-tuple based solution methodology for MOLOPs has also been proposed and used for student's performance evaluation in [11]. 2-tuple based solution methodology for MOLOPs has a serious limitation. It represents the semantics of linguistic information using type-1 (T1) FSs.

Interval type-2 (IT2) FSs capture and model the word uncertainty in a better manner than the T1 FSs [12], [13], [15]. Use of IT2 FSs for representing the word semantics in MOLOPs has been depicted in a recent work [16] (We will discuss the details in Section II). In [16], authors have proposed a novel perceptual reasoning (PR) based solution methodology for MOLOPs. PR is a novel design of CWW engine using if-then rules. In [17], PR based solution methodology has been used for car purchase problem, modelled as a LOP.

Therefore the aim of this work is to demonstrate the applicability of the PR based solution methodology for solving the MOLOP of parallel processor scheduling. We use the parallel processor scheduling problem of [9] and convert it to a MOLOP with two objectives viz., the overall completion time and overall profit. Our task here is to minimize the overall completion time and maximize overall profit. We have also compared the results obtained from application of PR based solution methodology to MOLOP of parallel processor scheduling, to those obtained with 2-tuple based solution methodology to the same MOLOP.

We have found the PR based solution methodology to be advantageous in comparison to the 2-tuple based methodology on three fronts. Firstly the PR based solution generates unique recommendations. Another advantage with PR based solution methodology is that the linguistic recommendation is a word from the codebook. Lastly, in PR based solution methodology, the word model comes before the word. All these advantages are not available with the 2-tuple based solution methodology. (We will discuss the details in Section V).

The organization of the rest of this paper is: Sections 2 gives the mathematical details of PR based solution methodology for MOLOPs, Section 3 illustrates the applicability of PR based solution methodology to MOLOP on parallel processor scheduling, Section 4 gives the mathematical details of 2-tuple based solution methodology for MOLOPs and illustrates the applicability of 2-tuple based solution methodology to MOLOP on parallel processor scheduling, Section 5 compares and discusses the results obtained with the PR and 2-tuple based solution methodologies, for MOLOP of parallel processor scheduling, and finally Section 6 concludes this paper as well as discusses its future scope.

## II. PERCEPTUAL REASONING BASED SOLUTION METHODOLOGY FOR MOLOPs

PR based solution methodology solves those MOLOPs, where the objective function/ functions and constraints are linked by if-then rules. This link is specified because the values of the objective function/ functions may not be known at all points of the decision space. A MOLOP in the generalized form may be given as:

$$max/min \{f_1(x), \dots f_K(x)\}$$
$$subject\ to\{ \Re_1(x), \dots \Re_n(x) | x \epsilon X \} \quad (1)$$

Here, $f_k, k = 1, \dots K$ are the objective functions, and $\Re_i$ is the $i^{th}$ if-then rule[2] described in the general form as follows:

$$R_i: IF\ x_1\ is\ \tilde{F}^i{}_1\ and \dots and\ IF\ x_n\ is\ \tilde{F}^i{}_n, THEN\ y_{1i}\ is\ \tilde{G}^i{}_1$$
$$and \dots and\ y_{qi}\ is\ \tilde{G}^i{}_q, i = 1, \dots, N \quad (2)$$

In (2), $x_j, j = 1, \dots, n$ are the antecedents, $y_{ki}, k = 1, \dots, q$ is the $k$th consequent of the $i^{th}$ rule. The antecedents and consequents take the values $\tilde{F}^i{}_j, j = 1, \dots, n$ and $\tilde{G}^i{}_k$, respectively. The $\tilde{F}^i{}_j$ and $\tilde{G}^i{}_k$, are linguistic values, whose semantics or word models are defined using IT2 FSs.

To generate the word models for $\tilde{F}^i{}_j j = 1, \dots, n$, the data about the interval end points of the words, can be collected by establishing a scale of 0 to 10. There are two methods to collect this data. In one method, a survey can be conducted among a group of subjects and they can be asked to provide the location of end points on the scale of 0 to 10. In another method called the Person FOU approach [23], a subject (or an expert) can be asked to provide an interval for each of the left and right end points. Then assuming uniform distribution, we generate 50 random numbers in left ($L_1, L_2, \dots, L_{50}$) and right intervals ($R_1, R_2, \dots, R_{50}$). Then 50 pairs $(L_i, R_i), i = 1\ to\ 50$, are formed so that each pair becomes a data interval, provided by $i^{th}$ virtual subject.

These data intervals are processed to generate the IT2 FS word models using either HMA [24], EIA [25] or IA [26], and stored in the codebook [15]. In this paper, we have used HMA as well as IA to generate IT2 FS word models for the codebook words. It is pertinent to mention that all the $\tilde{F}^i{}_j, j = 1, \dots, n$ are the IT2 FS word models from the codebook, whereas $\tilde{G}^i{}_k$, may not be from the codebook.

Let $\tilde{X}' = (\tilde{X}'_1, \dots, \tilde{X}'_N)$ be an $N \times 1$ vector of words input to each of the $N$ rules, given in generalized form of (2). Therefore, the firing level of $i^{th}$ rule, $f^i(\tilde{X}')$, is computed by performing the $minimum\ t - norm$ operation of the respective words from the input words vector and rule antecedents as:

$$f^i(\tilde{X}') = minimum\ t - norm(sm_j(\tilde{X}'_1, \tilde{F}^i{}_1), \dots, sm_j(\tilde{X}'_p, \tilde{F}^i{}_p))$$
$$(3)$$

where $sm_j(\tilde{X}'_j, \tilde{F}^i{}_j)$ is the Jaccard's similarity measure for IT2 FSs.

In PR, the fired rules are combined using LWA to generate the output IT2 FS for the PR corresponding to the $k$th consequent, $\tilde{Y}_{PR_k}$, given in (4) as:

$$\tilde{Y}_{PR_k} = \frac{\sum_{i=1,\dots,n; j=1,\dots,N} f^i(\tilde{X}') \tilde{G}^i{}_j}{\sum_{i=1,\dots,n} f^i(\tilde{X}')}, k = 1, \dots, q \quad (4)$$

---

[2] The if-then rules of (2) without MOLOP of (1) are similar to a fuzzy inference system, which employ a methodology to map a given input to an output using fuzzy logic. Examples of such methodologies are: the Mamdani inference method [18], [19], TSK method [20], [21] and Tsukamoto's inference method [22].

The condition on the $\tilde{Y}_{PR_k}, k = 1, \ldots, q$ from (4) is that it should be either interior, left shoulder or right shoulder FOU, and must resemble a codebook FOU. So, the computations as shown in (5)-(8) on the $\alpha$-cuts of $\tilde{G}^i_j$ and rule firing levels are performed as:

$$y_{Ll_k}(\alpha) = \min_{\forall f^i \in [\underline{f}^i, \overline{f}^i]} \frac{\sum_{i=1}^n a_{il}(\alpha) f^i}{\sum_{i=1}^n f^i}, \alpha \in [0,1], k = 1, \ldots, q \quad (5)$$

$$y_{Rr_k}(\alpha) = \max_{\forall f^i \in [\underline{f}^i, \overline{f}^i]} \frac{\sum_{i=1}^n b_{ir}(\alpha) f^i}{\sum_{i=1}^n f^i}, \alpha \in [0,1], k = 1, \ldots, q \quad (6)$$

$$y_{Lr_k}(\alpha) = \min_{\forall f^i \in [\underline{f}^i, \overline{f}^i]} \frac{\sum_{i=1}^n a_{ir}(\alpha) f^i}{\sum_{i=1}^n f^i}, \alpha \in [0, h_{\underline{Y}_{PR}}], k = 1, \ldots, q \quad (7)$$

$$y_{Rl_k}(\alpha) = \max_{\forall f^i \in [\underline{f}^i, \overline{f}^i]} \frac{\sum_{i=1}^n b_{il}(\alpha) f^i}{\sum_{i=1}^n f^i}, \alpha \in [0, h_{\underline{Y}_{PR}}], k = 1, \ldots, q \quad (8)$$

where the terms in (5)-(8) are as follows: the rule antecedents are expressed as lying inside the interval defined by the LMF and UMF, given by $\tilde{G}^i_j = [\underline{G}^i_j, \overline{G}^i_j]$, the height of LMF being $h_{\underline{G}^i_j}$, the $\alpha$-cut of $\underline{G}^i_j$ is denoted as $[a_{ir}(\alpha), b_{il}(\alpha)], \alpha \in [0, h_{\underline{G}^i_j}]$, the $\alpha$-cut of $\overline{G}^i_j$ is denoted as $[a_{il}(\alpha), b_{ir}(\alpha)], \alpha \in [0,1]$ and $h_{\underline{Y}_{PR}} = \min_i h_{\underline{G}^i_j}$. All these terms are depicted pictorially in Fig. 1. It is observed from Fig. 1 (a) that for left and right shoulder FOUs, $h_{\underline{G}^i_j} = 1$. Furthermore, for left shoulder FOU, $a_{il}(\alpha) = a_{ir}(\alpha) = 0 \ \forall \alpha \in [0,1]$ and right shoulder FOU, $b_{il}(\alpha) = b_{ir}(\alpha) = M \ \forall \alpha \in [0,1]$.

Using (5)-(8), the $\tilde{Y}_{PR_k}$ is given as: $\tilde{Y}_{PR_k} = [\underline{Y}_{PR_k}, \overline{Y}_{PR_k}]$, where $\underline{Y}_{PR_k}$ is completely characterized by $[y_{Lr_k}(\alpha), y_{Rl_k}(\alpha)]$ and $\overline{Y}_{PR_k}$ by $[y_{Ll_k}(\alpha), y_{Rr_k}(\alpha)]$. This is shown in Fig. 1 (b). Again it can be seen from Fig. 1 (b) that for left shoulder and right shoulder FOU, $y_{Ll_k}(\alpha) = y_{Lr_k}(\alpha) = 0 \ \forall \alpha \in [0,1]$ and $y_{Rl_k}(\alpha) = y_{Rr_k}(\alpha) = M \ \forall \alpha \in [0,1]$, respectively.

Finally a numeric recommendation corresponding to the $\tilde{Y}_{PR_k}$, is generated. We first use the Enhanced Karnik Mendel algorithm (EKM) [27] to generate the centroid end points, $c_l$ and $c_r$, for the IT2 FS word model corresponding to the $\tilde{Y}_{PR_k}$. Then we take the average of these centroid end points to generate a numeric recommendation corresponding to the $\tilde{Y}_{PR_k}$. To generate a linguistic recommendation for $\tilde{Y}_{PR_k}$, we compare its IT2 FS word model to those stored in the codebook, using Jaccard's similarity measure. The most similar word is recommended.

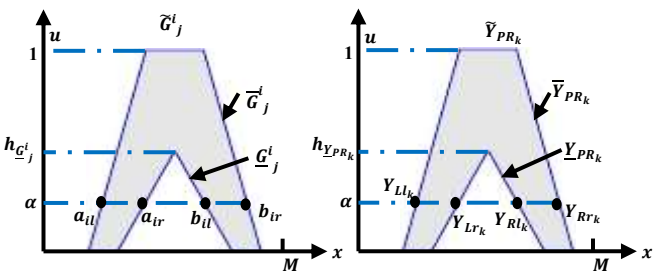

Fig. 1. (a) Words FOUs and the $\alpha$-cuts [5] (b) PR FOUs and the $\alpha$-cuts [5].

## III. PARALLEL PROCESSOR SCHEDULING: FORMULATION AS A MOLOP AND SOLUTION USING PR BASED METHODOLOGY

We consider the parallel processing scenario of [8], where a fixed number of welding jobs are to be accomplished by a given number of welders in the company. Let there be $n$ number of welding jobs and $k$ welders. Each welder welds the job but in a different amount of time. The processing time of each job is assessed on the basis of three criteria: welder ability ($WA$), batch size ($BS$) and welder's experience ($WE$). $WA$ is affected by a number of parameters such as training, etc. A welder, who learns the welding technique faster, can finish the same welding job earlier than the one who cannot learn the welding faster. $BS$, on the whole, denotes the total workload of jobs to be performed collectively or for individual welders, the number of jobs to be performed by him/ her.

Bigger the $BS$, more time is required to accomplish the task and vice versa. $WE$ is defined as the number of welding jobs performed by the welder in the past. If a welder has welded a job earlier, then he/ she can weld the same job faster, whenever it occurs the next time and vice versa. All these parameters take linguistic values, which are given in Table I. A number of assumptions are made such as any job can be performed by any welder; a welding job cannot be split and must be completed by one welder only, and pre-emptions are not allowed. Each welder required a certain amount of time to accomplish a task. This time was called the operation time ($OT$) of individual welders. The sum total of the $OTs$ of all the welders was called the overall completion time. The scenario in [8], consisted of only one objective viz., the minimization of overall completion time.

Table I
Parameters/ Outputs and Their Linguistic Values

| Parameters/ outputs | Linguistic values |
|---|---|
| Welder ability (WA) | Beginner (B) |
| | Slightly skilled (SS) |
| | Moderate (M) |
| | Good (G) |
| | Professional (P) |
| Batch size (BS) | Very small (VS) |
| | Small (S) |
| | Moderate (MS) |
| | Large (L) |
| | Extremely large (EL) |
| Welder experience (WE) | Very low (VL) |
| | Low (SLL) |
| | Moderate (SM) |
| | Large (SL) |
| | Very large (SVL) |
| Operation time (OT) | Very little (VLI) |
| | Small (SI) |
| | Moderate (MI) |
| | Large (LI) |
| | Very large (VLA) |
| Profit (PP) | Very less (VLP) |
| | Less (LP) |
| | Moderate (MP) |
| | High (H) |
| | Very high (VH) |

However, as this is seldom the case in real-life scenarios, therefore, we convert the scenario of [8] to a MOLOP, by adding another objective viz., the amount of profit ($PP$), derived from each of the individual welders. Thus, based on the values of $WA$, $WE$ and $BS$, the $OT$ of the individual welders as well as the $PP$ are computed. Therefore, the complete MOLOP is given as:

$$min/\max\{OT, PP\}$$
$$subject\ to\ wa_i \in WA, bs_i \in BS, we_i \in WE \quad (9)$$

In (9) $wa_i, bs_i$ and $we_i$ denote the welder's ability, batch size and welder's experience of the $i^{th}$ welder. Furthermore, $WA$, $BS$ and $WE$ denote here the term sets containing the linguistic terms corresponding to welder's ability, batch size and welder's experience, respectively. It is mentioned here that $WA$ and $WE$ show a different characteristic against the $OT$ as compared to $BS$. This happens because, more the $WE$, lesser time he/ she will take to weld the jobs and hence lesser $OT$. For the parameter $WA$, higher the welder's ability, lesser will be the $OT$. However, the smaller the $BS$, lesser will be the $OT$. The $OT$ shows a different behaviour against $PP$. As the overall completion time, is the sum total of $OTs$ of individual welders, therefore one objective of MOLOP in (9) is to minimize the $OT$ values for individual welders. Furthermore, since $OT$ shows a different behaviour against $PP$, therefore other objective is to maximize the value of $PP$.

Table II
End point intervals for linguistic terms used to rate the welders' parameters and outputs

| Parameters/ outputs | Linguistic term for Parameters/ outputs | Left end interval | | Right end interval | |
|---|---|---|---|---|---|
| | | Lower limit | Upper limit | Lower limit | Upper limit |
| Welder ability ($WA$) | Beginner ($B$) | 0 | 0 | 2 | 3 |
| | Slightly skilled ($SS$) | 0 | 0.5 | 4.5 | 5.5 |
| | Moderate ($M$) | 2 | 3 | 7 | 8 |
| | Good ($G$) | 4.5 | 5.5 | 9.5 | 10 |
| | Professional ($P$) | 7 | 8 | 10 | 10 |
| Batch size ($BS$) | Very small ($VS$) | 0 | 0 | 2 | 3 |
| | Small ($S$) | 0 | 0.5 | 4.5 | 5.5 |
| | Moderate ($MS$) | 2 | 3 | 7 | 8 |
| | Large ($L$) | 4.5 | 5.5 | 9.5 | 10 |
| | Extremely large ($EL$) | 7 | 8 | 10 | 10 |
| Welder experience ($WE$) | Very low ($VL$) | 0 | 0 | 2 | 3 |
| | Low ($SLL$) | 0 | 0.5 | 4.5 | 5.5 |
| | Moderate ($SM$) | 2 | 3 | 7 | 8 |
| | Large ($SL$) | 4.5 | 5.5 | 9.5 | 10 |
| | Very large ($SVL$) | 7 | 8 | 10 | 10 |
| Operation time ($OT$) | Very little ($VLI$) | 0 | 0 | 2 | 3 |
| | Small ($SI$) | 0 | 0.5 | 4.5 | 5.5 |
| | Moderate ($MI$) | 2 | 3 | 7 | 8 |
| | Large ($LI$) | 4.5 | 5.5 | 9.5 | 10 |
| | Very large ($VLA$) | 7 | 8 | 10 | 10 |
| Profit ($PP$) | Very less ($VLP$) | 0 | 0 | 2 | 3 |
| | Less ($LP$) | 0 | 0.5 | 4.5 | 5.5 |
| | Moderate ($MP$) | 2 | 3 | 7 | 8 |
| | High ($H$) | 4.5 | 5.5 | 9.5 | 10 |
| | Very high ($VH$) | 7 | 8 | 10 | 10 |

Table III
FOU words in the rule antecedents obtained with HMA. Each UMF and LMF is a trapezoid

| Parameters/ outputs | Linguistic term for Parameters/ outputs | UMF | | | | LMF | | | | | Centroid | | |
|---|---|---|---|---|---|---|---|---|---|---|---|---|---|
| | | | | | | | | | | | Left | Right | Mean |
| Welder ability ($WA$) | Beginner ($B$) | 0.00 | 0.00 | 2.00 | 3.54 | 0.00 | 0.00 | 2 | 3.26 | 1.00 | 1.35 | 1.43 | 1.39 |
| | Slightly skilled ($SS$) | 0.00 | 0.00 | 4.52 | 5.89 | 0.00 | 0.00 | 4.52 | 5.84 | 1.00 | 2.62 | 2.63 | 2.63 |
| | Moderate ($M$) | 1.00 | 2.96 | 7.02 | 9.03 | 1.81 | 2.96 | 7.02 | 8.27 | 1.00 | 4.8 | 5.23 | 5.01 |
| | Good ($G$) | 3.79 | 5.49 | 10.0 | 10.0 | 4.41 | 5.49 | 10.0 | 10.0 | 1.00 | 7.28 | 7.45 | 7.36 |
| | Professional ($P$) | 6.06 | 7.98 | 10.0 | 10.0 | 6.85 | 7.98 | 10.0 | 10.0 | 1.00 | 8.45 | 8.68 | 8.56 |
| Batch size ($BS$) | Very small ($VS$) | 0.00 | 0.00 | 2.07 | 3.87 | 0.00 | 0.00 | 2.07 | 3.32 | 1.00 | 1.38 | 1.54 | 1.46 |
| | Small ($S$) | 0.00 | 0.00 | 4.55 | 6.46 | 0.00 | 0.00 | 4.55 | 5.71 | 1.00 | 2.59 | 2.8 | 2.69 |
| | Moderate ($MS$) | 0.64 | 2.96 | 7.01 | 8.53 | 2 | 2.96 | 7.01 | 8.27 | 1.00 | 4.69 | 5.14 | 4.91 |
| | Large ($L$) | 3.53 | 5.48 | 10.0 | 10.0 | 4.4 | 5.48 | 10.0 | 10.0 | 1.00 | 7.21 | 7.44 | 7.32 |
| | Extremely large ($EL$) | 6.47 | 7.97 | 10.0 | 10.0 | 6.62 | 7.97 | 10.0 | 10.0 | 1.00 | 8.57 | 8.61 | 8.59 |
| Welder experience ($WE$) | Very low ($VL$) | 0.00 | 0.00 | 2.01 | 3.7 | 0.00 | 0.00 | 2.01 | 3.33 | 1.00 | 1.37 | 1.48 | 1.43 |
| | Low ($SLL$) | 0.00 | 0.00 | 4.53 | 6.56 | 0.00 | 0.00 | 4.53 | 5.8 | 1.00 | 2.61 | 2.82 | 2.72 |
| | Moderate ($SM$) | 1.20 | 2.96 | 7.02 | 8.98 | 1.79 | 2.96 | 7.02 | 8.19 | 1.00 | 4.83 | 5.2 | 5.02 |
| | Large ($SL$) | 3.84 | 5.45 | 10.0 | 10.0 | 4.29 | 5.45 | 10.0 | 10.0 | 1.00 | 7.29 | 7.41 | 7.35 |
| | Very large ($SVL$) | 6.08 | 7.97 | 10.0 | 10.0 | 6.75 | 7.97 | 10.0 | 10.0 | 1.00 | 8.45 | 8.65 | 8.55 |
| Operation time ($OT$) | Very little ($VLI$) | 0.00 | 0.00 | 2.01 | 4.06 | 0.00 | 0.00 | 2.01 | 3.09 | 1.00 | 1.31 | 1.59 | 1.45 |
| | Small ($SI$) | 0.00 | 0.00 | 4.52 | 6.31 | 0.00 | 0.00 | 4.52 | 5.85 | 1.00 | 2.62 | 2.75 | 2.69 |
| | Moderate ($MI$) | 1.40 | 2.98 | 7.02 | 8.56 | 1.79 | 2.98 | 7.02 | 8.22 | 1.00 | 4.9 | 5.1 | 5.00 |
| | Large ($LI$) | 3.8 | 5.49 | 10.0 | 10.0 | 4.12 | 5.49 | 10.0 | 10.0 | 1.00 | 7.29 | 7.37 | 7.33 |
| | Very large ($VLA$) | 6.29 | 7.99 | 10.0 | 10.0 | 6.77 | 7.99 | 10.0 | 10.0 | 1.00 | 8.52 | 8.66 | 8.59 |
| Profit ($PP$) | Very less ($VLP$) | 0.00 | 0.00 | 2.04 | 3.75 | 0.00 | 0.00 | 2.04 | 3.26 | 1.00 | 1.36 | 1.5 | 1.43 |
| | Less ($LP$) | 0.00 | 0.00 | 4.51 | 6.2 | 0.00 | 0.00 | 4.51 | 5.7 | 1.00 | 2.58 | 2.72 | 2.65 |
| | Moderate ($MP$) | 1.11 | 2.98 | 7.02 | 8.97 | 1.65 | 2.98 | 7.02 | 8.12 | 1.00 | 4.79 | 5.17 | 4.98 |
| | High ($H$) | 3.75 | 5.46 | 10.0 | 10.0 | 4.21 | 5.46 | 10.0 | 10.0 | 1.00 | 7.26 | 7.39 | 7.33 |
| | Very high ($VH$) | 6.05 | 7.99 | 10.0 | 10.0 | 6.76 | 7.99 | 10.0 | 10.0 | 1.00 | 8.45 | 8.65 | 8.55 |

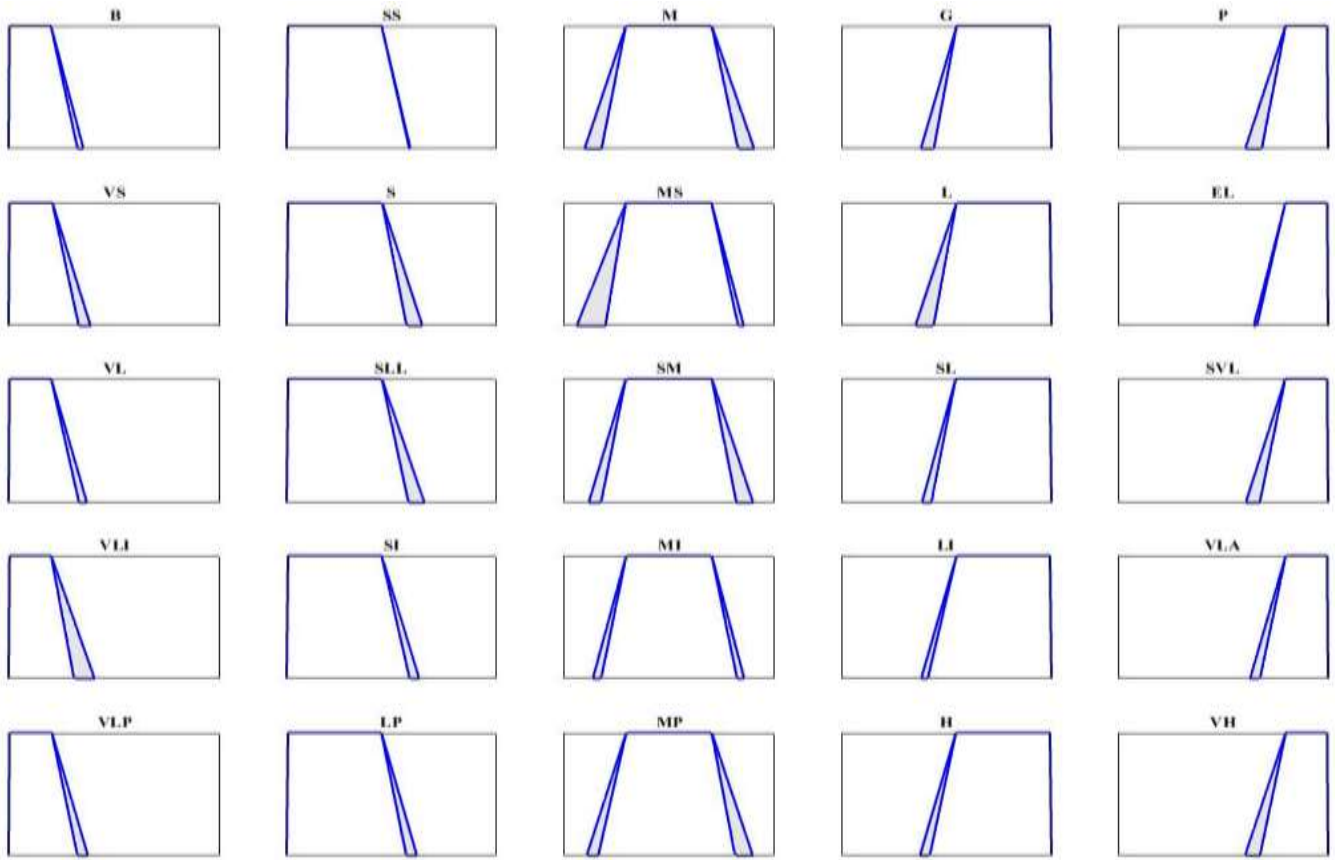

Fig. 2 Codebook used in parallel processor scheduling generated using HMA

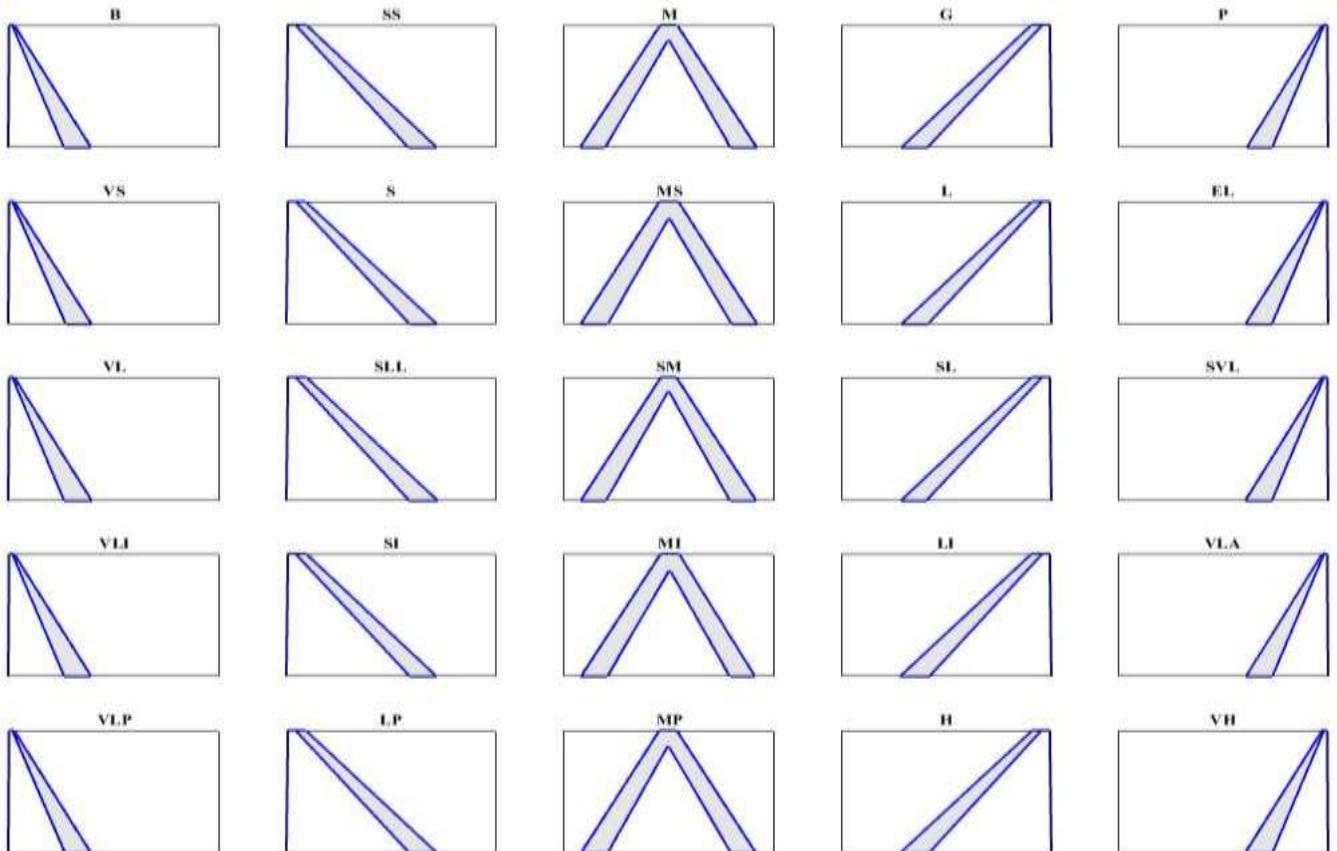

Fig. 3 Codebook used in parallel processor scheduling generated using IA

Table IV
FOU words in the rule antecedents obtained with IA. Each UMF and LMF is a trapezoid

| Parameters/ outputs | Linguistic term for parameters/ outputs | UMF | | | | LMF | | | | Centroid | | |
|---|---|---|---|---|---|---|---|---|---|---|---|---|
| | | | | | | | | | | Left | Right | Mean |
| Welder ability (WA) | Beginner (B) | 0.00 | 0.00 | 0.27 | 3.91 | 0.00 | 0.00 | 0.18 | 2.63 | 1.00 | 0.88 | 1.35 | 1.12 |
| | Slightly skilled (SS) | 0.00 | 0.00 | 0.94 | 7.16 | 0.00 | 0.00 | 0.43 | 5.81 | 1.00 | 1.94 | 2.49 | 2.22 |
| | Moderate (M) | 0.8 | 4.6 | 5.39 | 9.16 | 1.99 | 4.99 | 4.99 | 7.92 | 0.88 | 4.43 | 5.52 | 4.98 |
| | Good (G) | 2.86 | 9.06 | 10.0 | 10.0 | 4.11 | 9.58 | 10.0 | 10.0 | 1.00 | 7.52 | 8.03 | 7.78 |
| | Professional (P) | 6.13 | 9.73 | 10.0 | 10.0 | 7.34 | 9.81 | 10.0 | 10.0 | 1.00 | 8.67 | 9.11 | 8.89 |
| Batch size (BS) | Very small (VS) | 0.00 | 0.00 | 0.27 | 3.94 | 0.00 | 0.00 | 0.19 | 2.73 | 1.00 | 0.92 | 1.35 | 1.13 |
| | Small (S) | 0.00 | 0.00 | 0.91 | 7.17 | 0.00 | 0.00 | 0.45 | 5.87 | 1.00 | 1.97 | 2.49 | 2.23 |
| | Moderate (MS) | 0.83 | 4.57 | 5.45 | 9.18 | 2.11 | 5.01 | 5.01 | 7.97 | 0.87 | 4.45 | 5.59 | 5.02 |
| | Large (L) | 2.86 | 9.09 | 10.0 | 10.0 | 4.13 | 9.58 | 10.0 | 10.0 | 1.00 | 7.52 | 8.04 | 7.78 |
| | Extremely large (EL) | 6.06 | 9.73 | 10.0 | 10.0 | 7.33 | 9.81 | 10.0 | 10.0 | 1.00 | 8.65 | 9.1 | 8.87 |
| Welder experience (WE) | Very low (VL) | 0.00 | 0.00 | 0.28 | 3.95 | 0.00 | 0.00 | 0.18 | 2.65 | 1.00 | 0.89 | 1.36 | 1.12 |
| | Low (SLL) | 0.00 | 0.00 | 0.93 | 7.21 | 0.00 | 0.00 | 0.44 | 5.86 | 1.00 | 1.96 | 2.5 | 2.23 |
| | Moderate (SM) | 0.83 | 4.64 | 5.35 | 9.14 | 2.02 | 4.99 | 4.99 | 7.89 | 0.89 | 4.45 | 5.51 | 4.98 |
| | Large (SL) | 2.83 | 9.09 | 10.0 | 10.0 | 4.04 | 9.57 | 10.0 | 10.0 | 1.00 | 7.51 | 8.01 | 7.76 |
| | Very large (SVL) | 6.05 | 9.72 | 10.0 | 10.0 | 7.33 | 9.81 | 10.0 | 10.0 | 1.00 | 8.64 | 9.1 | 8.87 |
| Operation time (OT) | Very little (VLI) | 0.00 | 0.00 | 0.27 | 3.92 | 0.00 | 0.00 | 0.18 | 2.65 | 1.00 | 0.89 | 1.35 | 1.12 |
| | Small (SI) | 0.00 | 0.00 | 0.94 | 7.14 | 0.00 | 0.00 | 0.43 | 5.89 | 1.00 | 1.97 | 2.48 | 2.22 |
| | Moderate (MI) | 0.85 | 4.62 | 5.48 | 9.07 | 2.1 | 5.04 | 5.04 | 7.88 | 0.87 | 4.44 | 5.55 | 5 |
| | Large (LI) | 2.8 | 9.07 | 10.0 | 10.0 | 4.2 | 9.58 | 10.0 | 10.0 | 1.00 | 7.5 | 8.06 | 7.78 |
| | Very large (VLA) | 6.09 | 9.73 | 10.0 | 10.0 | 7.35 | 9.82 | 10.0 | 10.0 | 1.00 | 8.65 | 9.11 | 8.88 |
| Profit (PP) | Very less (VLP) | 0.00 | 0.00 | 0.27 | 3.91 | 0.00 | 0.00 | 0.19 | 2.69 | 1.00 | 0.9 | 1.34 | 1.12 |
| | Less (LP) | 0.00 | 0.00 | 0.92 | 7.17 | 0.00 | 0.00 | 0.44 | 5.86 | 1.00 | 1.96 | 2.49 | 2.22 |
| | Moderate (MP) | 0.85 | 4.56 | 5.37 | 9.12 | 2.11 | 4.97 | 4.97 | 7.94 | 0.88 | 4.45 | 5.54 | 4.99 |
| | High (H) | 2.8 | 9.1 | 10.0 | 10.0 | 4.18 | 9.54 | 10.0 | 10.0 | 1.00 | 7.51 | 8.05 | 7.78 |
| | Very high (VH) | 6.05 | 9.72 | 10.0 | 10.0 | 7.35 | 9.82 | 10.0 | 10.0 | 1.00 | 8.64 | 9.11 | 8.88 |

Table V
FOU data for rule consequent words obtained with HMA. Each UMF and LMF is a trapezoid

| Output | Consequent word | UMF | | | | LMF | | | | Centroid | | |
|---|---|---|---|---|---|---|---|---|---|---|---|---|
| | | | | | | | | | | Left | Right | Mean |
| Operation time (OT) | $\tilde{G}^1_{OT}$ | 0.00 | 0.00 | 2.01 | 4.06 | 0.00 | 0.00 | 2.01 | 3.09 | 1.00 | 1.31 | 1.59 | 1.45 |
| | $\tilde{G}^2_{OT}$ | 0.00 | 0.00 | 4.52 | 6.31 | 0.00 | 0.00 | 4.52 | 5.85 | 1.00 | 2.62 | 2.75 | 2.69 |
| | $\tilde{G}^3_{OT}$ | 1.4 | 2.98 | 7.02 | 8.56 | 1.79 | 2.98 | 7.02 | 8.22 | 1.00 | 4.9 | 5.1 | 5 |
| | $\tilde{G}^4_{OT}$ | 3.8 | 5.49 | 10.0 | 10.0 | 4.12 | 5.49 | 10.0 | 10.0 | 1.00 | 7.29 | 7.37 | 7.33 |
| | $\tilde{G}^5_{OT}$ | 6.29 | 7.99 | 10.0 | 10.0 | 6.77 | 7.99 | 10.0 | 10.0 | 1.00 | 8.52 | 8.66 | 8.59 |
| Profit (PP) | $\tilde{G}^1_{PP}$ | 6.05 | 7.99 | 10.0 | 10.0 | 6.76 | 7.99 | 10.0 | 10.0 | 1.00 | 8.45 | 8.65 | 8.55 |
| | $\tilde{G}^2_{PP}$ | 3.75 | 5.46 | 10.0 | 10.0 | 4.21 | 5.46 | 10.0 | 10.0 | 1.00 | 7.26 | 7.39 | 7.33 |
| | $\tilde{G}^3_{PP}$ | 1.11 | 2.98 | 7.02 | 8.97 | 1.65 | 2.98 | 7.02 | 8.12 | 1.00 | 4.79 | 5.17 | 4.98 |
| | $\tilde{G}^4_{PP}$ | 0.00 | 0.00 | 4.51 | 6.2 | 0.00 | 0.00 | 4.51 | 5.7 | 1.00 | 2.58 | 2.72 | 2.65 |
| | $\tilde{G}^5_{PP}$ | 0.00 | 0.00 | 2.04 | 3.75 | 0.00 | 0.00 | 2.04 | 3.26 | 1.00 | 1.36 | 1.5 | 1.43 |

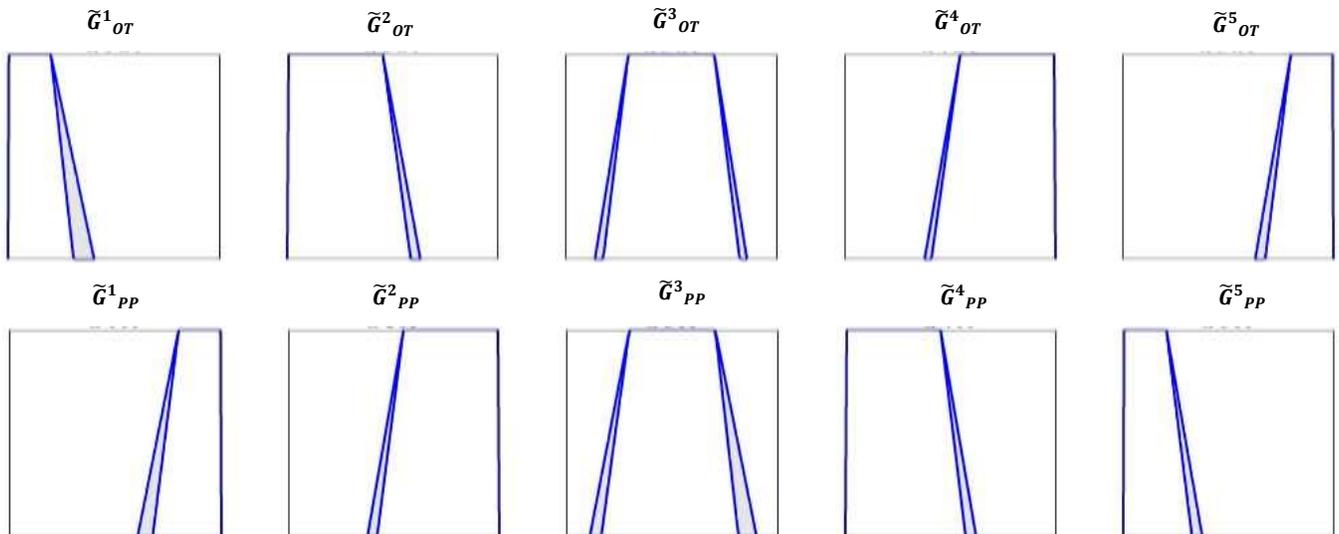

Fig. 4 Rule consequents generated using HMA

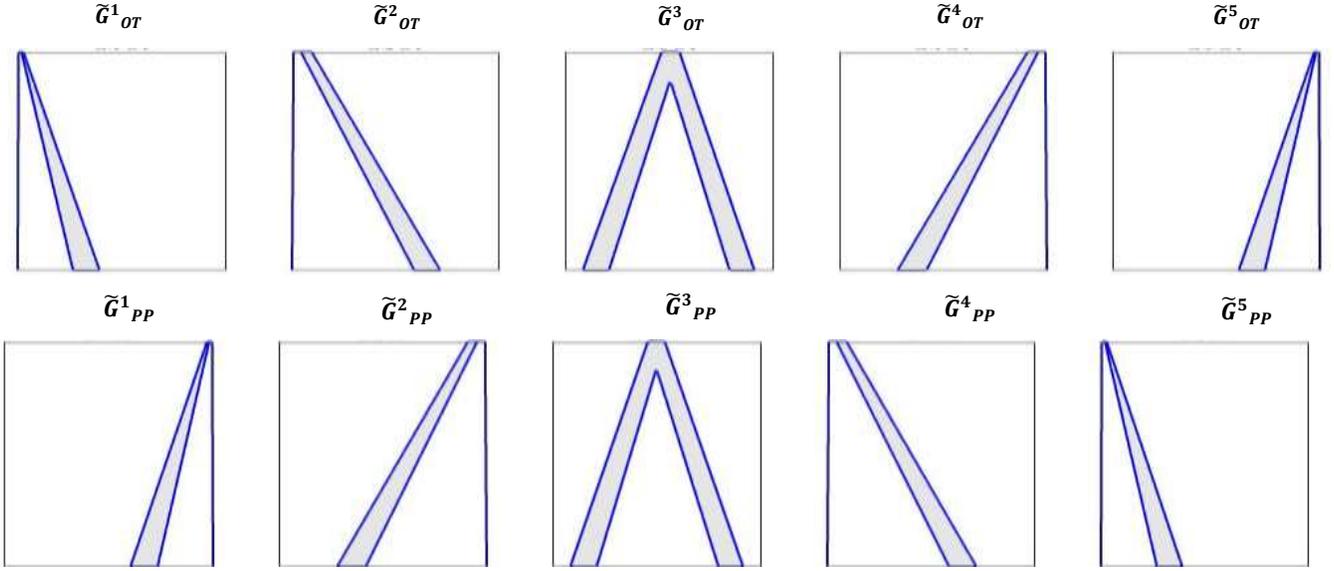

Fig. 5 Rule consequents generated using IA

Table VI
FOU data for rule consequent words obtained with IA. Each UMF and LMF is a trapezoid

| Output | Consequent word | UMF | | | | LMF | | | | Centroid | | |
|---|---|---|---|---|---|---|---|---|---|---|---|---|
| | | | | | | | | | | Left | Right | Mean |
| Operation time (OT) | $\tilde{G}^1_{OT}$ | 0.00 | 0.00 | 0.27 | 3.92 | 0.00 | 0.00 | 0.18 | 2.65 | 1.00 | 0.89 | 1.35 | 1.12 |
| | $\tilde{G}^2_{OT}$ | 0.00 | 0.00 | 0.94 | 7.14 | 0.00 | 0.00 | 0.43 | 5.89 | 1.00 | 1.97 | 2.48 | 2.22 |
| | $\tilde{G}^3_{OT}$ | 0.85 | 4.62 | 5.48 | 9.07 | 2.1 | 5.04 | 5.04 | 7.88 | 0.87 | 4.44 | 5.55 | 5 |
| | $\tilde{G}^4_{OT}$ | 2.8 | 9.07 | 10.0 | 10.0 | 4.2 | 9.58 | 10.0 | 10.0 | 1.00 | 7.5 | 8.06 | 7.78 |
| | $\tilde{G}^5_{OT}$ | 6.09 | 9.73 | 10.0 | 10.0 | 7.35 | 9.82 | 10.0 | 10.0 | 1.00 | 8.65 | 9.11 | 8.88 |
| Profit (PP) | $\tilde{G}^1_{PP}$ | 6.05 | 9.72 | 10.0 | 10.0 | 7.35 | 9.82 | 10.0 | 10.0 | 1.00 | 8.64 | 9.11 | 8.88 |
| | $\tilde{G}^2_{PP}$ | 2.8 | 9.1 | 10.0 | 10.0 | 4.18 | 9.54 | 10.0 | 10.0 | 1.00 | 7.51 | 8.05 | 7.78 |
| | $\tilde{G}^3_{PP}$ | 0.85 | 4.56 | 5.37 | 9.12 | 2.11 | 4.97 | 4.97 | 7.94 | 0.88 | 4.45 | 5.54 | 4.99 |
| | $\tilde{G}^4_{PP}$ | 0.00 | 0.00 | 0.92 | 7.17 | 0.00 | 0.00 | 0.44 | 5.86 | 1.00 | 1.96 | 2.49 | 2.22 |
| | $\tilde{G}^5_{PP}$ | 0.00 | 0.00 | 0.27 | 3.91 | 0.00 | 0.00 | 0.19 | 2.69 | 1.00 | 0.9 | 1.34 | 1.12 |

Table VII
Welder parameter values and firing levels computed with HMA and IA

| Welder | Parameters | | | Firing levels | | | | | | | | | |
|---|---|---|---|---|---|---|---|---|---|---|---|---|---|
| | WA | BS | WE | HMA | | | | | IA | | | | |
| | | | | $f^1$ | $f^2$ | $f^3$ | $f^4$ | $f^5$ | $f^1$ | $f^2$ | $f^3$ | $f^4$ | $f^5$ |
| 1 | P | VS | SM | 0.10 | 0.39 | 0.1 | 0.001 | 0.10 | 0.067 | 0.006 | 0.065 | 0.005 | 0.067 |
| 2 | G | MS | SVL | 0.11 | 0.41 | 0.1 | 0.002 | 0.11 | 0.069 | 0.245 | 0.067 | 0.006 | 0.069 |
| 3 | M | MS | SVL | 0.101 | 0.39 | 0.1 | 0.002 | 0.101 | 0.065 | 0.244 | 0.067 | 0.006 | 0.065 |
| 4 | SS | L | SM | 0.001 | 0.059 | 0.381 | 0.402 | 0.001 | 0.005 | 0.066 | 0.248 | 0.25 | 0.005 |
| 5 | B | MS | SM | 0.001 | 0.001 | 0.089 | 0.381 | 0.001 | 0.001 | 0.005 | 0.069 | 0.248 | 0.001 |

Now we present the applicability of PR based solution methodology to the MOLOP of parallel processor scheduling, discussed in Section II. The solution methodology begins by deciding the linguistic terms of the term set from which the antecedents and rule consequents take values. Then the end point intervals of these linguistic terms are located on the scale of 0 to 10 and using Person FOU approach, a codebook is generated. We decide the interval end points for the linguistic terms of the parameters as well as the outputs on the scale of 0 to 10, and list them in the Table II. Based on the Person FOU approach, the codebook is generated using HMA and IA, and is shown in Fig. 2 and Fig. 3, respectively. We show the codebook data generated using HMA in Table III and using IA in Table IV.

Let the system be designed using five if-then rules given in (10) as:

$\Re_1$: if WA is P and BS is VS and WE is SVL then $\tilde{G}^1_{OT}$ is VLI and $\tilde{G}^1_{PP}$ is VH

$\Re_2$: if WA is G and BS is S and WE is SL then $\tilde{G}^2_{OT}$ is SI and $\tilde{G}^2_{PP}$ is H

$\Re_3$: if WA is M and BS is MS and WE is SM then $\tilde{G}^3_{OT}$ is MI and $\tilde{G}^3_{PP}$ is MP

$\Re_4$: if WA is SS and BS is L and WE is SLL then $\tilde{G}^4_{OT}$ is LI and $\tilde{G}^4_{PP}$ is LP

$\Re_5$: if WA is B and BS is EL and WE is VL then $\tilde{G}^5_{OT}$ is VLA and $\tilde{G}^5_{PP}$ is VLP   (10)

In (10), $\tilde{G}^i{}_{OT}$ $i = 1,..,5$ and $\tilde{G}^i{}_{PP}, i = 1,..,5$ correspond to the consequent linguistic value for $OT$ and $PP$, respectively. Therefore, the linguistic terms for the rule consequents are also represented by IT2 FSs. Their FOU plots generated using HMA and IA is shown in Fig. 4 and 5, respectively. The corresponding FOU data is given in Table V and VI, respectively.

Now let's consider a scenario where the values of the respective parameters of welders are given in Table VII (rows 4 to 8 and columns 2 to 4). Thus, using the PR based solution methodology for MOLOP (given in Section II), the respective rules are fired and the firing levels are computed. For example, consider the data values of welder 1. The word vector corresponding to his/ her values for antecedent terms input to rule $\Re_1$ is $(\tilde{X}'_1 = P, \tilde{X}'_2 = VS, \tilde{X}'_3 = SM)$. Thus, firing level computation is performed as shown in (11) as:

$$f^1 = min\ t - norm\left(sm_j(\tilde{X}'_1, P), sm_j(\tilde{X}'_2, VS), sm_j(\tilde{X}'_3, SVL)\right)$$
$$= min\ t - norm\left(sm_j(P, P), sm_j(VS, VS), sm_j(SM, SVL)\right)$$
$$= min\ t - norm\ (1,1,0.1) = 0.1 \qquad (11)$$

Similarly, other firing levels are computed with HMA and IA. These values are listed in Table VIII.

Now we aggregate the linguistic values in the rule consequents to arrive at the values of the objective functions, similar to (4). Thus, the aggregated values of the two objective functions viz., $OT$ and $PP$ (given in (9)) for $i^{th}$ welder are obtained as:

$$OT_i = \frac{f^1\tilde{G}^1{}_{OT} + f^2\tilde{G}^2{}_{OT} + f^3\tilde{G}^3{}_{OT} + f^4\tilde{G}^4{}_{OT} + f^5\tilde{G}^5{}_{OT}}{f^1 + f^2 + f^3 + f^4 + f^5}, i = 1,...,5 \qquad (12)$$

$$PP_i = \frac{f^1\tilde{G}^1{}_{PP} + f^2\tilde{G}^2{}_{PP} + f^3\tilde{G}^3{}_{PP} + f^4\tilde{G}^4{}_{PP} + f^5\tilde{G}^5{}_{PP}}{f^1 + f^2 + f^3 + f^4 + f^5}\ i = 1,...,5 \qquad (13)$$

where the quantities in (12) and (13) are as follows: $OT_i, i = 1,...,5$ and $PP_i, i = 1,...,5$ is the objective function for operation time and profit, respectively of $i^{th}$ welder, $f^i$, $\tilde{G}^i{}_{OT}$ and $\tilde{G}^i{}_{PP}$ $i = 1,...,5$ is the firing level value, linguistic value of consequent for the operation time and linguistic value of consequent for the profit, respectively obtained from $i^{th}$ rule. Therefore, for welder 1, the values of his/ her firing levels obtained with HMA are given in row 4 and columns 2 to 4 of Table V. Thus, the corresponding values of his/ her objective functions are given in (14-(15) as:

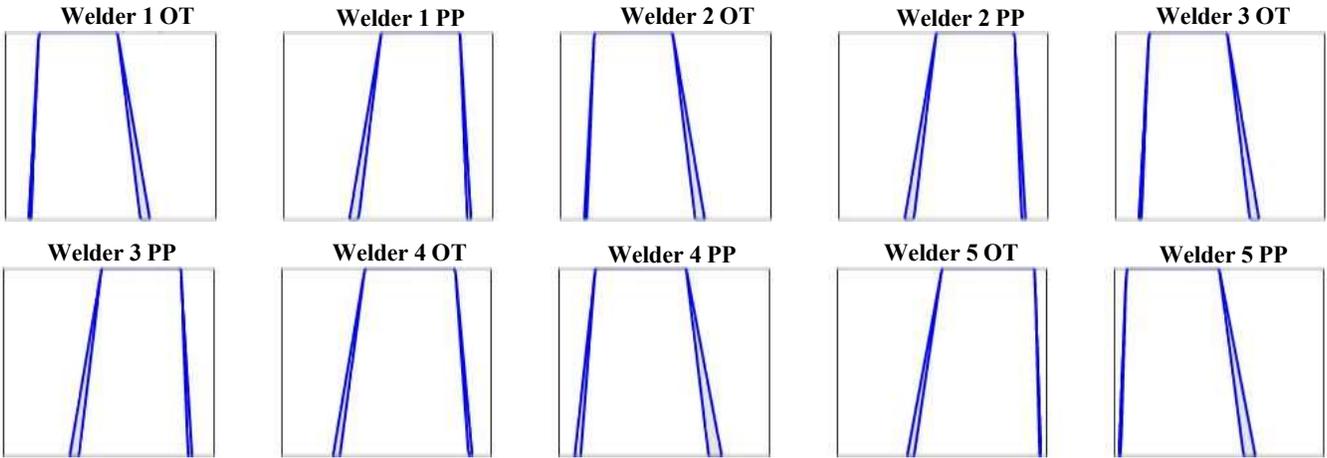

Fig. 6 Welder objective function plots generated using HMA

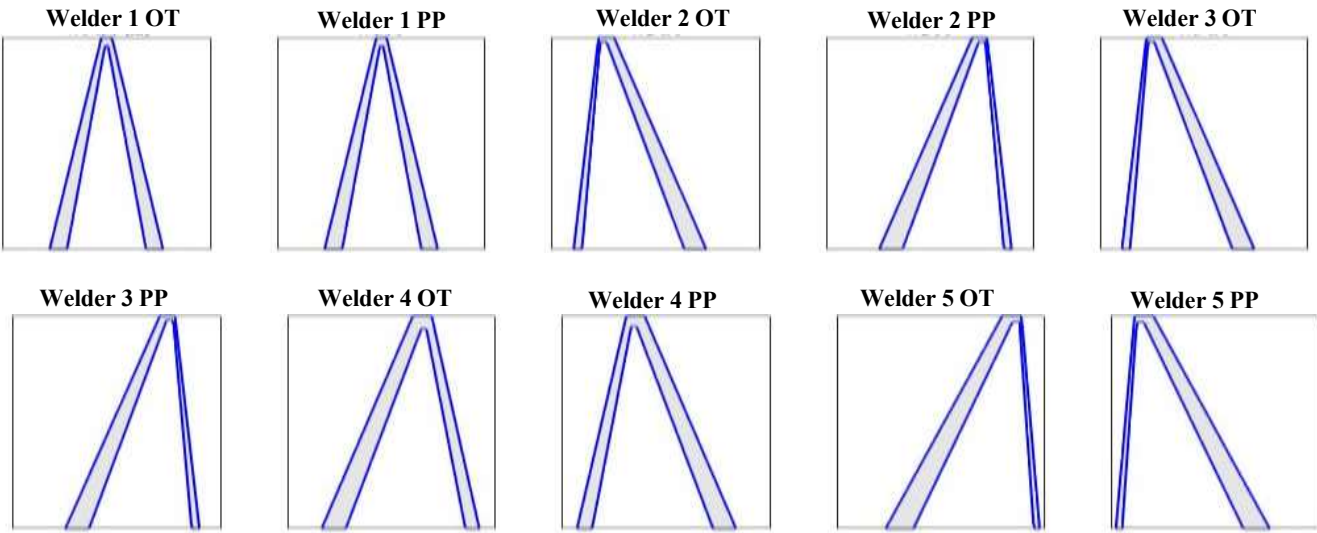

Fig. 7 Welder objective function plots generated using IA

Table VIII
FOU data for objective functions obtained with HMA. Each UMF and LMF is a trapezoid

| Welder | Objective function | UMF | | | | LMF | | | | Centroid | | | Linguistic recommendation |
|---|---|---|---|---|---|---|---|---|---|---|---|---|---|
| | | | | | | | | | | Left | Right | Mean | |
| 1 | Operation time (OT) | 1.12 | 1.6 | 5.32 | 6.85 | 1.24 | 1.6 | 5.32 | 6.4 | 1.00 | 3.62 | 3.78 | 3.71 | Small (SI) |
| | Profit (PP) | 3.15 | 4.67 | 8.41 | 8.94 | 3.59 | 4.67 | 8.41 | 8.75 | 1.00 | 6.21 | 6.39 | 6.31 | High (H) |
| 2 | Operation time (OT) | 1.15 | 1.62 | 5.32 | 6.84 | 1.27 | 1.62 | 5.32 | 6.39 | 1.00 | 3.63 | 3.8 | 3.72 | Small (SI) |
| | Profit (PP) | 3.16 | 4.67 | 8.38 | 8.91 | 3.6 | 4.67 | 8.38 | 8.72 | 1.00 | 6.2 | 6.38 | 6.29 | High (H) |
| 3 | Operation time (OT) | 1.13 | 1.61 | 5.33 | 6.85 | 1.26 | 1.61 | 5.33 | 6.41 | 1.00 | 3.63 | 3.79 | 3.71 | Small (SI) |
| | Profit (PP) | 3.15 | 4.66 | 8.4 | 8.93 | 3.59 | 4.66 | 8.4 | 8.74 | 1.00 | 6.21 | 6.38 | 6.3 | High (H) |
| 4 | Operation time (OT) | 2.45 | 3.97 | 8.26 | 9.08 | 2.78 | 3.97 | 8.26 | 8.9 | 1.00 | 5.88 | 6.02 | 5.95 | Moderate (MI) |
| | Profit (PP) | 0.77 | 1.74 | 6.03 | 7.72 | 1.05 | 1.74 | 6.03 | 7.1 | 1.00 | 3.91 | 4.16 | 4.03 | Moderate (MP) |
| 5 | Operation time (OT) | 3.34 | 5 | 9.41 | 9.71 | 3.67 | 5 | 9.41 | 9.64 | 1.00 | 6.82 | 6.92 | 6.87 | Large (LI) |
| | Profit (PP) | 0.23 | 0.59 | 5 | 6.73 | 0.33 | 0.59 | 5 | 6.17 | 1.00 | 3.02 | 3.2 | 3.11 | Less (LP) |

Table IX
FOU data for objective functions obtained with IA. Each UMF and LMF is a trapezoid

| Welder | Objective function | UMF | | | | LMF | | | | Centroid | | | Linguistic recommendation |
|---|---|---|---|---|---|---|---|---|---|---|---|---|---|
| | | | | | | | | | | Left | Right | Mean | |
| 1 | Operation time (OT) | 2.27 | 4.75 | 5.24 | 7.69 | 3.1 | 4.92 | 5.06 | 6.88 | 0.96 | 4.65 | 5.32 | 4.99 | Moderate (MI) |
| | Profit (PP) | 2.27 | 4.77 | 5.25 | 7.72 | 3.12 | 4.94 | 5.09 | 6.93 | 0.96 | 4.68 | 5.34 | 5.01 | Moderate (MP) |
| 2 | Operation time (OT) | 1.08 | 2.27 | 3 | 7.41 | 1.48 | 2.35 | 2.64 | 6.37 | 0.98 | 3.25 | 3.84 | 3.54 | Small (SI) |
| | Profit (PP) | 2.54 | 7.03 | 7.73 | 8.91 | 3.67 | 7.34 | 7.65 | 8.54 | 0.98 | 6.16 | 6.75 | 6.46 | High (H) |
| 3 | Operation time (OT) | 1.05 | 2.23 | 2.96 | 7.42 | 1.44 | 2.31 | 2.6 | 6.37 | 0.98 | 3.23 | 3.81 | 3.52 | Small (SI) |
| | Profit (PP) | 2.54 | 7.06 | 7.77 | 8.94 | 3.67 | 7.38 | 7.69 | 8.57 | 0.98 | 6.18 | 6.78 | 6.48 | High (H) |
| 4 | Operation time (OT) | 1.64 | 6.03 | 6.92 | 9.22 | 2.8 | 6.44 | 6.67 | 8.55 | 0.94 | 5.49 | 6.28 | 5.89 | Moderate (MI) |
| | Profit (PP) | 0.74 | 3.1 | 3.96 | 8.33 | 1.46 | 3.33 | 3.58 | 7.24 | 0.95 | 3.72 | 4.49 | 4.1 | Moderate (MP) |
| 5 | Operation time (OT) | 2.34 | 7.96 | 8.87 | 9.74 | 3.68 | 8.44 | 8.77 | 9.46 | 0.97 | 6.75 | 7.42 | 7.09 | Large (LI) |
| | Profit (PP) | 0.24 | 1.14 | 2.03 | 7.63 | 0.54 | 1.24 | 1.58 | 6.37 | 0.97 | 2.59 | 3.24 | 2.91 | Less (LP) |

Table X
FOU data for overall completion time and profit obtained with HMA and IA. Each UMF and LMF is a trapezoid

| Technique | Objective function | UMF | | | | LMF | | | | Centroid | | | Linguistic recommendation |
|---|---|---|---|---|---|---|---|---|---|---|---|---|---|
| | | | | | | | | | | Left | Right | Mean | |
| HMA | Overall completion time | 1.84 | 2.76 | 6.73 | 7.87 | 2.04 | 2.76 | 6.73 | 7.55 | 1 | 4.72 | 4.86 | 4.79 | Moderate (MI) |
| | Overall profit | 2.09 | 3.27 | 7.24 | 8.25 | 2.43 | 3.27 | 7.24 | 7.9 | 1 | 5.11 | 5.3 | 5.21 | Moderate (MP) |
| IA | Overall completion time | 1.68 | 4.65 | 5.4 | 8.3 | 2.5 | 4.89 | 5.15 | 7.53 | 0.97 | 4.67 | 5.33 | 5.01 | Moderate (MI) |
| | Overall profit | 1.67 | 4.62 | 5.35 | 8.31 | 2.49 | 4.85 | 5.12 | 7.53 | 0.97 | 4.67 | 5.32 | 4.99 | Moderate (MP) |

Table XI
Welder parameter values and firing levels

| Welder | Parameters | | | Firing levels | Operation time indices | Profit indices | Overall completion time | | Overall profit | |
|---|---|---|---|---|---|---|---|---|---|---|
| | WA | BS | WE | | | | Numeric | Linguistic | Numeric | Linguistic |
| 1 | P | VS | SM | $\alpha_1 = 15$ | $p_{C^1{}_{OT}} = 1,$ | $p_{C^1{}_{PP}} = 5,$ | | | | |
| 2 | G | MS | SVL | $\alpha_2 = 60$ | $p_{C^2{}_{OT}} = 2,$ | $p_{C^2{}_{PP}} = 4,$ | | | | |
| 3 | M | MS | SVL | $\alpha_3 = 45$ | $p_{C^3{}_{OT}} = 3,$ | $p_{C^3{}_{PP}} = 3,$ | 2.69 | (MI, −0.31) | 3.31 | (MP, 0.31) |
| 4 | SS | L | SM | $\alpha_4 = 24$ | $p_{C^4{}_{OT}} = 4,$ | $p_{C^4{}_{PP}} = 2,$ | | | | |
| 5 | B | MS | SM | $\alpha_5 = 9$ | $p_{C^5{}_{OT}} = 5$ | $p_{C^5{}_{PP}} = 1$ | | | | |

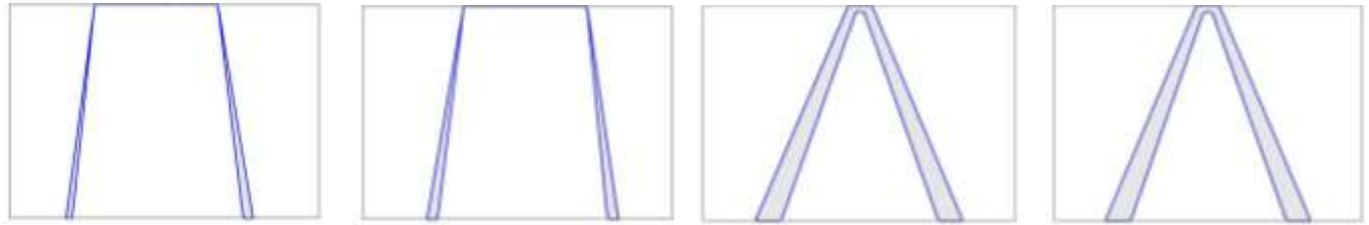

(a) Overall completion time with HMA     (b) Overall profit with HMA     (c) Overall completion time with IA     (b) Overall profit with IA

Fig. 8 FOU plots for overall completion time and profit obtained with HMA and IA

$$OT_1 = \frac{0.1 \times \widetilde{G}^1{}_{OT} + 0.39 \times \widetilde{G}^2{}_{OT} + 0.1 \times \widetilde{G}^3{}_{OT} + 0.001 \times \widetilde{G}^4{}_{OT} + 0.1 \times \widetilde{G}^5{}_{OT}}{0.1 + 0.39 + 0.1 + 0.001 + 0.1} \quad (14)$$

$$PP_1 = \frac{0.1 \times \widetilde{G}^1{}_{PP} + 0.39 \times \widetilde{G}^2{}_{PP} + 0.1 \times \widetilde{G}^3{}_{PP} + 0.001 \times \widetilde{G}^4{}_{PP} + 0.1 \times \widetilde{G}^5{}_{PP}}{0.1 + 0.39 + 0.1 + 0.001 + 0.1} \quad (15)$$

The values of the linguistic variables used in the rule consequents obtained with HMA viz., $\tilde{G}^i_{OT}$ and $\tilde{G}^i_{PP}$ $i = 1, \ldots, 5$ are given in Table V. Thus, the result of (14) and (15) is the operation time and profit, respectively for welder 1, obtained with HMA.

Similarly, the values of operation time and profit can be obtained for other welders as well. Furthermore, substituting the values of firing levels and rule consequents obtained with IA, gives the value of the two objective functions with IA. The values obtained with HMA are summarized in Table VIII and with IA in Table IX. The corresponding FOU plots are given in Fig. 6 and 7, respectively.

To calculate the values of overall completion time ($C_{overall}$) and profit ($P_{overall}$), the values of these objective for respective welders can be aggregated using the LWA as:

$$C_{overall} = \frac{\sum_{i=1}^{5} \widetilde{OT}_i \widetilde{W}_i}{\sum_{i=1}^{5} \widetilde{W}_i} \quad (16)$$

$$P_{overall} = \frac{\sum_{i=1}^{5} \widetilde{PP}_i \widetilde{W}_i}{\sum_{i=1}^{5} \widetilde{W}_i} \quad (17)$$

In (16) and (17), the quantities are: $\widetilde{OT}_i$ is the operation time and $\widetilde{PP}_i$ profit of $i^{th}$ welder obtained from (12) and (13), respectively. Furthermore, $\widetilde{W}_i$ denotes the importance of $i^{th}$ welder. We assume all welders to be equally important here. It is mentioned here that a different notation used in (16) and (17) for operation time and profit, respectively denotes that operation time and profit are represented in the form of IT2 FSs. Thus, using (16) and (17), the values of overall completion time and profit, respectively with HMA and IA are obtained. These values are shown in Table X and corresponding FOU plots in Fig. 8.

IV. 2-TUPLE BASED SOLUTION METHODOLOGY FOR MOLOPs: MATHEMATICAL PRELIMINARIES AND SOLUTION OF MOLOP ON PARALLEL PROCESSOR SCHEDULING

We now present the mathematical preliminaries of 2-tuple linguistic model based solution methodology for MOLOPs and then illustrate it applicability to the MOLOP on parallel processor scheduling, from Section III.

*A. 2-Tuple Based Solution Methodologies For MOLOPS: Mathematical preliminaries*

Consider a MOLOP with if-then rules given in (1). With the 2-tuple based solution methodology for MOLOPs, the $i^{th}$ if-then rule $\Re_i$ is described in the general form as follows:

$$\Re_i(x): if\ x_1\ is\ A_{i1}\ and\ \ldots x_n\ is\ A_{in}\ then\ f_1(x)\ is\ C_{i1} \ldots and\ f_K(x)\ is\ C_{iK} \quad (18)$$

Let the total number of if-then rules be $N$. The first step in the 2-tuple based solution methodology is to identify the linguistic variables used in the antecedent and consequent parts of if-then rules corresponding to the MOLOP and define their collections as respective term sets. The linguistic variables used in the antecedent part of MOLOP of (18) are: $A_{i1}, \ldots, A_{in}, \ldots, A_{N1}, \ldots, A_{Nn}$. Corresponding values of linguistic variables in the consequent part are: $C_{i1}, \ldots, C_{iK}, \ldots, C_{N1}, \ldots, C_{NK}$. Therefore, the term set corresponding to the linguistic variables in the antecedent and consequent part of MOLOP is given as (19) and (20), respectively:

$$S_1 = \{s_p : A_{ij} |\ p = n \times (i-1) + j; i = 1, \ldots, N; j = 1, \ldots, n\} \quad (19)$$
$$S_2 = \{s_q : C_{ij} |\ q = K \times (i-1) + j; i = 1, \ldots, N; j = 1, \ldots, K\} \quad (20)$$

In (19) and (20), $p$ and $q$, respectively is the index of the $ij^{th}$ linguistic term from antecedent and consequent part, respectively of the $i^{th}$ if-then rule.

Next step in the 2-tuple based solution methodology for MOLOPs, is the computation of firing level values for each of the $i^{th}$ if-then rule $\Re_i$, using $product\ t - norm$, and thus the resulting firing level is called the $i^{th}$ firing level. During the calculation of the $i^{th}$ firing level value, we multiply the indices of the linguistic terms occurring in antecedent part of the $i^{th}$ if-then rule. For example, in rule $\Re_i$, the antecedent values are given as: $A_{i1}, \ldots, A_{in}$. Their respective indices according to the term set of (19) are: $n \times (i-1) + 1, \ldots, n \times (i-1) + n$. Thus, the product of these terms to give the value of $i^{th}$ firing level is given as:

$$\alpha_i = [n \times (i-1) + 1] \times \ldots \times [n \times (i-1) + n]; i = 1, \ldots, N \quad (21)$$

Thus, the firing level values are computed for all the $N$ rules. Now, the values of the linguistic variables used in the rule consequents are shown in (20). In (20), the value of the $l^{th}$ objective in $i^{th}$ rule are given as $C_{ij}$. Therefore, using (22), these values of the indices of the objective functions and firing levels are combined.

$$f_k(y) \coloneqq \frac{\alpha_1 C_{1i}(\alpha_1) + \ldots \ldots \alpha_m C_{mi}(\alpha_m)}{\alpha_1 + \ldots \ldots \alpha_m}, l = 1, \ldots, N \quad (22)$$

Finally, each of these computed function values are converted into the 2-tuple form by performing computations as shown in (23)-(25) as:

$$\beta_l = f_l \quad (23)$$
$$r = round\ (\beta_l) \quad (24)$$
$$\alpha_l = \beta_l - r \quad (25)$$

Thus, the recommended output function value using (23)-(25) is given as $(s_{round(\beta_l)}, \alpha_l)$.

*B. MOLOP on Parallel processor scheduling: solution using 2-tuple based solution methodology*

The linguistic terms corresponding to various parameters and output, used in MOLOP of parallel processor scheduling, are given in Table I. Their semantics are represented as triangular MFs in Fig. 9, on scale of 0 to 1 [28]. We define term sets corresponding to linguistic terms used in rule antecedents and consequents, for use in 2-tuple based solution methodology for MOLOPs. These term sets are given as:

$$WA = \{s_1 : B, s_2 : SS, s_3 : M, s_4 : G, s_5 : P\}$$
$$BS = \{s_1 : VS, s_2 : S, s_3 : MS, s_4 : L, s_5 : EL\}$$
$$WE = \{s_1 : VL, s_2 : SLL, s_3 : SM, s_4 : SL, s_5 : SVL\}$$
$$OT = \{s_1 : VLI, s_2 : SI, s_3 : MI, s_4 : LI, s_5 : VLA\}$$
$$PP = \{s_1 : VLP, s_2 : LP, s_3 : MP, s_4 : H, s_5 : VH\} \quad (26)$$

The if-then rules used to design the system are given in (10). The notations of the if-then rules are modified for 2-tuple representation and shown in (27) as:

$\Re_1$: if $WA$ is $P$ and $BS$ is $VS$ and $WE$ is $SVL$ then $C^1_{OT}$ is $VLI$ and $C^1_{PP}$ is $VH$

$\Re_2$: if $WA$ is $G$ and $BS$ is $S$ and $WE$ is $SL$ then $C^2_{OT}$ is $SI$ and $C^2_{PP}$ is $H$

$\Re_3$: if $WA$ is $M$ and $BS$ is $MS$ and $WE$ is $SM$ then $C^2_{OT}$ is $MI$ and $C^3_{PP}$ is $MP$

$\Re_4$: if $WA$ is $SS$ and $BS$ is $L$ and $WE$ is $SLL$ then $C^4_{OT}$ is $LI$ and $C^4_{PP}$ is $LP$

$\Re_5$: if $WA$ is $B$ and $BS$ is $EL$ and $WE$ is $VL$ then $C^5_{OT}$ is $VLA$ and $C^5_{PP}$ is $VLP$ (27)

Values of various parameters for the five welders are given in Table VIII (rows 4 to 8 and columns 2 to 4). For processing them using 2-tuple based solution methodology, we first compute the firing levels. Consider the data of first welder for finding out the value of first firing level. He/ she is a professional ($P$), has very small ($VS$) batch size to weld and moderate ($SM$) amount of experience. The indices corresponding to the linguistic values of these parameters are extracted from (26) and are found to be: 5, 1 and 3, respectively. Thus, the firing level is computed as:

$$\alpha_1 = 5 \times 1 \times 3 = 15 \quad (28)$$

Similarly, values of other firing levels viz., $\alpha_i, i = 2, \ldots, 5$, are also computed. These values are summarized in Table XI. Based on the values of firing levels and linguistic terms used in rule consequents, the overall completion time ($C_{overall}$) and profit ($P_{overall}$) are computed as shown in (29) and (30), respectively.

$$C_{overll} = \frac{\alpha_1 \times p_{C^1_{OT}} + \alpha_2 \times p_{C^2_{OT}} + \alpha_3 \times p_{C^3_{OT}} + \alpha_4 \times p_{C^4_{OT}} + \alpha_5 \times p_{C^5_{OT}}}{\alpha_1 + \alpha_2 + \alpha_3 + \alpha_4 + \alpha_5}$$
(29)

$$P_{overll} = \frac{\alpha_1 \times p_{C^1_{PP}} + \alpha_2 \times p_{C^2_{PP}} + \alpha_3 \times p_{C^3_{PP}} + \alpha_4 \times p_{C^4_{PP}} + \alpha_5 \times p_{C^5_{PP}}}{\alpha_1 + \alpha_2 + \alpha_3 + \alpha_4 + \alpha_5}$$
(30)

In (29) $p_{C^i_{OT}}, i = 1, \ldots, 5$ is the index of the linguistic term extracted from term set $OT$ of (26), corresponding to operation time from $i^{th}$ rule consequent of (27). In (30), $p_{C^i_{PP}}, i = 1, \ldots, 5$ is the corresponding index for profit. These values are also given in Table XI. Thus, using (29) and (30), we get:

$$C_{overll} = \frac{15 \times 1 + 60 \times 2 + 45 \times 3 + 24 \times 4 + 9 \times 5}{15 + 60 + 45 + 24 + 9}$$
$$= 2.69 \quad (31)$$

$$P_{overll} = \frac{15 \times 5 + 60 \times 4 + 45 \times 3 + 24 \times 2 + 9 \times 1}{15 + 60 + 45 + 24 + 9}$$
$$= 33.8 \quad (32)$$

The overall computation time in (31) is converted to 2-tuple form by performing following computations as:

$$\beta = 2.69 \quad (33)$$
$$j = round\ (2.69) = 3 \quad (34)$$
$$\alpha = \beta - round\ (\beta) = -0.31 \quad (35)$$

Therefore, the recommended linguistic term is given as: $(s_j, \alpha) = (s_3, -0.31) = (MI, -0.31)$. Thus, the $OT$ value for this operator is *Moderate*. Similarly, the overall profit is given as: $(MP, 0.31)$.

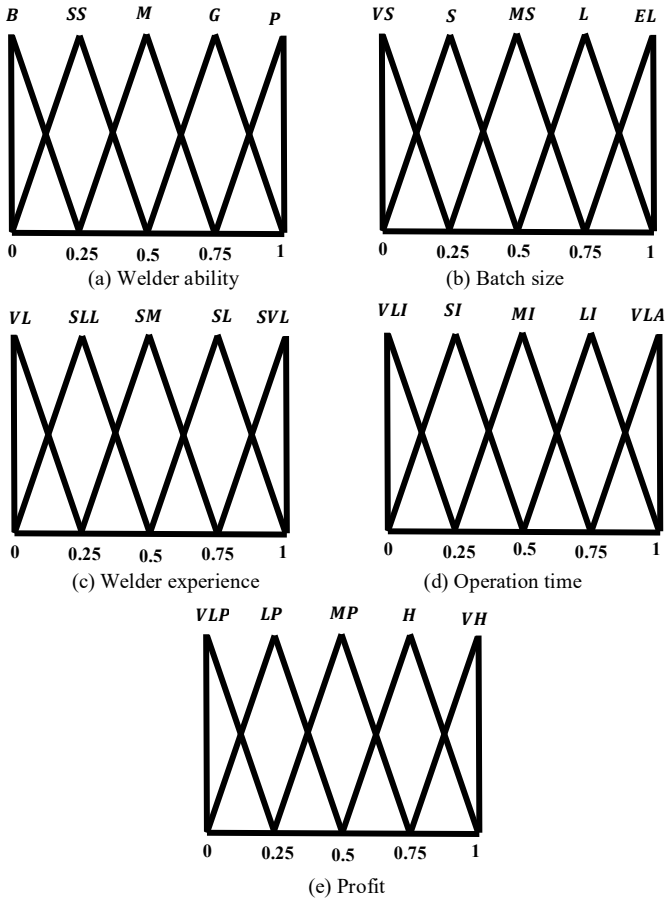

Fig 9 Membership function representation of linguistic terms of different parameters and the output

Table XII
Welders' Operation times and profits as well as Overall completion time and Overall profits obtained with PR and 2-tuple based solution methodologies

| Welder | Operation time | | | | | | Profit | | | | | | Overall completion time | | | | | | Overall profit | | | | | |
|---|---|---|---|---|---|---|---|---|---|---|---|---|---|---|---|---|---|---|---|---|---|---|---|---|
| | HMA | | IA | | 2-tuple | | HMA | | IA | | 2-tuple | | HMA | | IA | | 2-tuple | | HMA | | IA | | 2-tuple | |
| | N[a] | L[b] | N[a] | L[b] | N[a] | L[b] | N[a] | L[b] | N[a] | L[b] | N[a] | L[b] | N[a] | L[b] | N[a] | L[b] | N[a] | L[b] | N[a] | L[b] | N[a] | L[b] | N[a] | L[b] |
| 1 | 3.71 | SI | 4.99 | MI | 1 | VLI | 6.31 | H | 5.01 | MP | 5 | VH | | | | | | | | | | | | |
| 2 | 3.72 | SI | 3.54 | SI | 2 | SI | 6.29 | H | 6.46 | H | 4 | H | | | | | | | | | | | | |
| 3 | 3.71 | SI | 3.52 | SI | 3 | MI | 6.30 | H | 6.48 | H | 3 | MP | 4.79 | MI | 5.01 | MI | 2.69 | (MI,−0.31) | 5.21 | MP | 4.99 | MP | 3.31 | (MP,0.31) |
| 4 | 5.95 | MI | 5.89 | MI | 4 | LI | 4.03 | MP | 4.10 | MP | 2 | LP | | | | | | | | | | | | |
| 5 | 6.87 | LI | 7.09 | LI | 5 | VLA | 3.11 | LP | 2.91 | LP | 1 | VLP | | | | | | | | | | | | |

[a]Numeric, [b]Linguistic

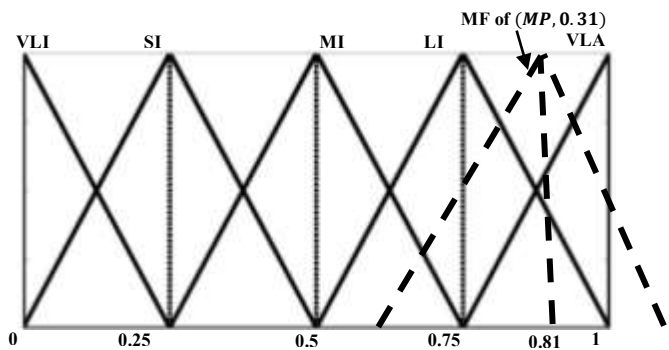

Fig. 10 Overall profit obtained with 2-tuple based solution methodology

## V. DISCUSSIONS

In the scenario of parallel processor scheduling, we considered parameter values for five welders. We calculated their individual operations times and profits. Furthermore, using these individual values, we calculated the overall completion time and overall profit. With PR, we used two techniques viz., HMA and IA. In the Table XII, we present the results obtained with PR based solution methodology (HMA and IA), as well as 2-tuple based solution methodology. From Table XII, it can be seen that the linguistic recommendations for operation time and profit for individual welders are generated using HMA and IA are the same (except first welder). Also, these recommendations are different from those obtained with 2-tuple based solution methodology.

We feel that the PR based solution methodology is better than 2-tuple based solution methodology based on three major aspects.

1) Firstly, PR based solution methodology can uniquely differentiate between the recommendations, by a combination of linguistic and numeric consideration. For example, consider the operation times for welders 2 and 3 (rows 5 and 6, column 2). Though both have same value of operation time as $Small$ ($SI$), however, both differ in numeric values viz., 3.71 and 3.72. On the other hand, with 2-tuple approach, if a scenario had existed where two welders had same values of operation times as $Small$ ($SI$), then both would had the numeric value as 2 for the operation times, thus rendering the two indistinguishable.

2) Another improvement of PR based solution methodology over 2-tuple based methodology is that the linguistic recommendation generated by former is a word from the codebook, whereas by the latter is generally not from the codebook. For example, consider the linguistic values of overall profit obtained with PR based solution methodology through HMA and IA, given in columns 21 and 23, respectively of Table XII. These values are $Moderate$ ($MP$) in both cases. The FOU plots of these linguistic values are shown in Fig. 8, and they match and FOU plot from the codebook, given in the Fig. 2 (for HMA) and Fig. 3 (for IA). However, the linguistic recommendation obtained with 2-tuple approach is ($MP$, 0.31) (Please see column 25 of Table XII). The semantics of the recommendation ($MP$, 0.31), is a T1 MF, at a distance of 0.31 from T1 MF of MP, as shown in Fig. 10. Thus, the recommendation obtained from 2-tuple based methodology may not be an exact match to a codebook word.

3) Another key aspect is the way the if-then rules are designed with PR based solution methodology and the 2-tuple based methodology. For example, in if-then rules of (10), corresponding to the PR based solution methodology, the word models for rule consequents can be constructed using the feedback of people [29]. However, in if-then rules of (27), corresponding to the 2-tuple based solution methodology, the word models for rule consequents must be constructed apriori. Thus, the PR based solution methodology is a CWW in true sense where, word model comes before the word, which is not the case with 2-tuple based solution methodology [15].

## VI. CONCLUSIONS AND FUTURE SCOPE

The emphasis of this work is on the parallel processor scheduling, which forms the backbone of the Industry 4.0. Industry 4.0 may be interpreted as the future industrial design mechanism which aims to minimize the involvement of human beings and maximizing the automation in various industrial processes. But, we strongly feel that how so ever minimum the involvement of human beings in the industrial processes, the primacy of the experts cannot be overlooked. In fact the use of expert opinions helps to design robust and performance oriented systems. As experts are generally human beings, with vast knowledge pertaining to the system under design, therefore, they tend to express their opinions naturally using linguistic terms or words. These linguistic opinions pertain to various scheduling criteria, which enable the optimal allocation of resources to attain the optimal value of an objective.

In real-life scenarios, the parallel processor scheduling seldom have a single objective. Also, FSs are used to model the word semantics in a manner quite close to the human cognitive process. Furthermore, IT2 FSs are better at word modelling than the T1 FSs. Thus, all these factors collectively motivated us to model the parallel processor scheduling scenario as a MOLOP, and solve it using the novel PR based solution methodology.

We have also compared the results obtained with the application of PR based solution methodology to the MOLOP of parallel processor scheduling to those obtained with 2-tuple based solution methodology. We have demonstrated that the PR based solution methodology gives a better performance than the 2-tuple based solution methodology on three fronts.

Firstly the PR based solution generates unique recommendations (by a combination of numeric and linguistic recommendations). Secondly, the linguistic recommendation generated by this solution methodology, is a word whose semantics are represented in the form of IT2 FSs in the codebook. Finally, in PR based solution methodology, the word model comes before the word. The 2-tuple based solution methodology falls short to offer these advantages and thus render itself as limiting with application to MOLOPs.

In future, the work maybe extended to MOLOPs, where there are more than two objectives, as well as they are differentially weighted.

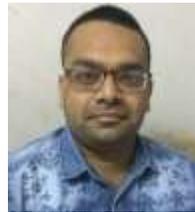

**Prashant K Gupta** (M'14) received the B.Tech and M.Tech degrees from the Guru Gobind Singh Indraprastha University, New Delhi, India in 2008 and 2012, respectively. He has completed his Ph.D degree in Computer Science from South Asian University, New Delhi, India. His research interests include fuzzy logic, computing with words, linguistic optimization and energy management. Prashant has published in reputed journals/conferences such as IEEE Transactions on Fuzzy Systems, Fuzzy Sets and Systems, Applied Soft Computing, Granular Computing, Fuzz- IEEE and IEEE SMC.

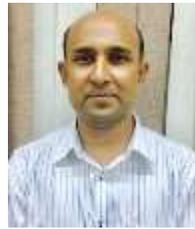

**Pranab K. Muhuri** (M'08) was born in Chittagong, Bangladesh. He received his Ph.D. degree in Computer Engineering in 2005 from IT-BHU [now Indian Institute of Technology, (BHU)], Varanasi, India. He is currently a Professor with the Department of Computer Science, South Asian University, New Delhi, India, where he is leading the computational intelligence research group. His current research interests are mainly in real-time systems, computational intelligence, especially fuzzy systems, evolutionary algorithms, perceptual computing, and machine learning. Pranab is an active member of the IEEE Computer Society, IEEE Computational Intelligence Society and Association of Computing Machinery (ACM). Pranab has published about 60 research papers in well-known journals and conferences including IEEE Transactions on Fuzzy Systems, Fuzzy Sets and Systems, Applied Soft Computing, Future Generation Computing Systems, Computers and Industrial Engineering, and Engineering Applications of Artificial Intelligence. Currently, he is serving as a member of the Editorial Board of the Applied Soft Computing journal.